\documentclass[runningheads]{llncs}
\usepackage{graphicx}
\usepackage{comment}
\usepackage{amsmath,amssymb} %
\usepackage{color}
\usepackage{subfigure}
\usepackage{enumitem}
\usepackage{booktabs}
\usepackage{adjustbox}
\usepackage{multirow}
\usepackage[breaklinks=true,bookmarks=false]{hyperref}

\usepackage{xspace}
\makeatletter
\DeclareRobustCommand\onedot{\futurelet\@let@token\@onedot}
\def\@onedot{\ifx\@let@token.\else.\null\fi\xspace}

\def\eg{\emph{e.g}\onedot}

\def\etal{\emph{et al}\onedot}
\makeatother

\usepackage{xcolor}

\newcommand{\yf}[1]{#1}

\newcommand{\bx}{\mathbf{x}}
\newcommand{\bt}{\mathbf{t}}
\newcommand{\bs}{\mathbf{s}}

\begin{document}
\pagestyle{headings}
\mainmatter
\def\ECCVSubNumber{676}  %

\title{Unifying Specialist Image Embedding into Universal Image Embedding} %

\titlerunning{Universal Image Embedding}
\author{Yang Feng\inst{1}\thanks{Part of this work was done while Yang Feng was an Intern in Google.} \and Futang Peng\inst{2} \and Xu Zhang\inst{3} \and
Wei Zhu\inst{1} \and Shanfeng Zhang\inst{2} \and Howard Zhou\inst{2} \and
Zhen Li\inst{2} \and Tom Duerig\inst{2} \and Shih-Fu Chang\inst{3} \and
Jiebo Luo\inst{1}
}
\authorrunning{Y. Feng et al.}
\institute{University of Rochester \and Google Research \and Columbia University}
\maketitle

\begin{abstract}
Deep image embedding provides a way to measure the semantic similarity of two images. It plays a central role in many applications such as image search, face verification, and zero-shot learning. It is desirable to have a universal deep embedding model applicable to various domains of images. \yf{However, existing methods mainly rely on training specialist embedding models each of which is applicable to images from a single domain.} In this paper, we study an important but unexplored task: how to train a single universal image embedding model to match the performance of several specialists on each specialist's domain. Simply fusing the training data from multiple domains cannot solve this problem because some domains become overfitted sooner when trained together using existing methods. Therefore, we propose to distill the knowledge in multiple specialists into a universal embedding to solve this problem. In contrast to existing embedding distillation methods that distill the absolute distances between images, we transform the absolute distances between images into a probabilistic distribution and minimize the KL-divergence between the distributions of the specialists and the universal embedding. Using several public datasets, we validate that our proposed method accomplishes the goal of universal image embedding.

\end{abstract}

\section{Introduction}
A fundamental problem in computer vision is how to measure the similarity or distance between a pair of images. With a good distance metric for images, it is possible to perform clustering~\cite{xing2003distance}, retrieval ~\cite{oh2016deep}, and classification~\cite{weinberger2009distance}, as well as verifying whether two images are from the same category~\cite{schroff2015facenet}. This problem has been studied for decades but not been fully solved yet.

The emergence of the deep image embedding has greatly advanced the performance of distance involving tasks. With the deep convolutional neural networks~(CNNs)~\cite{lecun1998gradient}, both low-level and high-level features are learned directly from raw images instead of being hand-crafted by human experts. The recent research in deep image embedding mainly focuses on how to design a better loss function~\cite{qian2019softtriple,sohn2016improved}, sampling policy~\cite{duan2019deep,wu2017sampling}, or proxy-based method~\cite{movshovitz2017no}.

\yf{Existing deep image embedding approaches generally consider images from one domain as input each time.} For example, for fine-grained car image retrieval, car images with different brands and models are collected and a car specialist embedding is trained with all the car images. The resulted car embedding may work well for car images but not for images from a different domain, \eg birds. For fine-grained bird image retrieval, the same process needs to be repeated. 

\begin{figure}[t]
  \centering
  \includegraphics[width=\columnwidth]{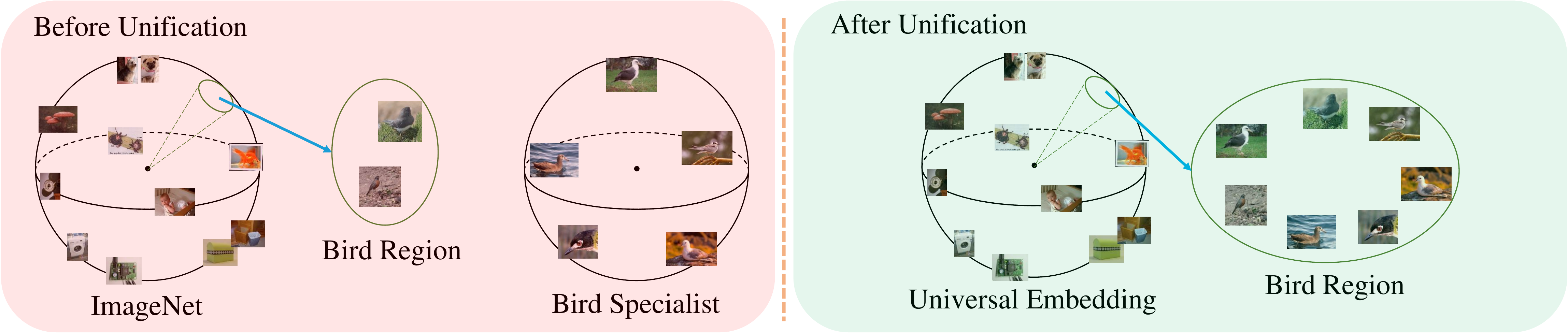}
  \setlength{\abovecaptionskip}{-0.4cm} 
  \caption{Before unifying the ImageNet specialist and the bird specialist, the embedding performance of the ImageNet specialist on bird images is lower than that of the bird specialist. We aim to use the bird specialist to enrich or strengthen the bird sub-space inside the ImageNet specialist. After unification, the performance of the universal embedding model on bird images should be similar to the bird specialist performance. %
  }
  \label{fig:1}
\vspace{-0.7cm}
\end{figure}

\yf{Modern content-based image search engines need to deal with arbitrary query images uploaded by a user. To achieve good retrieval results in practice, several specialist embedding models are maintained. When the query image is detected as belonging to one specialist's domain, that specialist is used for retrieval. In addition, a default embedding model is also needed when a query image does not belong to any of the trained specialist models' domain.} Keeping one image embedding model for each domain has three drawbacks. First, it is needed to decide which embedding model to use for a given query image. If the car embedding model is used for extracting the embedding vector for a bird image, the extracted embedding vector cannot reflect the discriminative information of the bird image. Choosing the correct embedding model for query images alone is a difficult task. \yf{It is impossible to achieve perfect precision or recall over unrestricted query images uploaded by users. Second, multiple embedding vectors need to be computed and stored for one gallery image, which takes significantly large storage in large-scale image search. Third, it is impractical to distribute so many embedding models to mobile devices.}

\begin{figure}[t]
\vspace{-2mm}
  \centering
\subfigure{\label{fig:overfitting}
  \includegraphics[width=0.28\columnwidth]{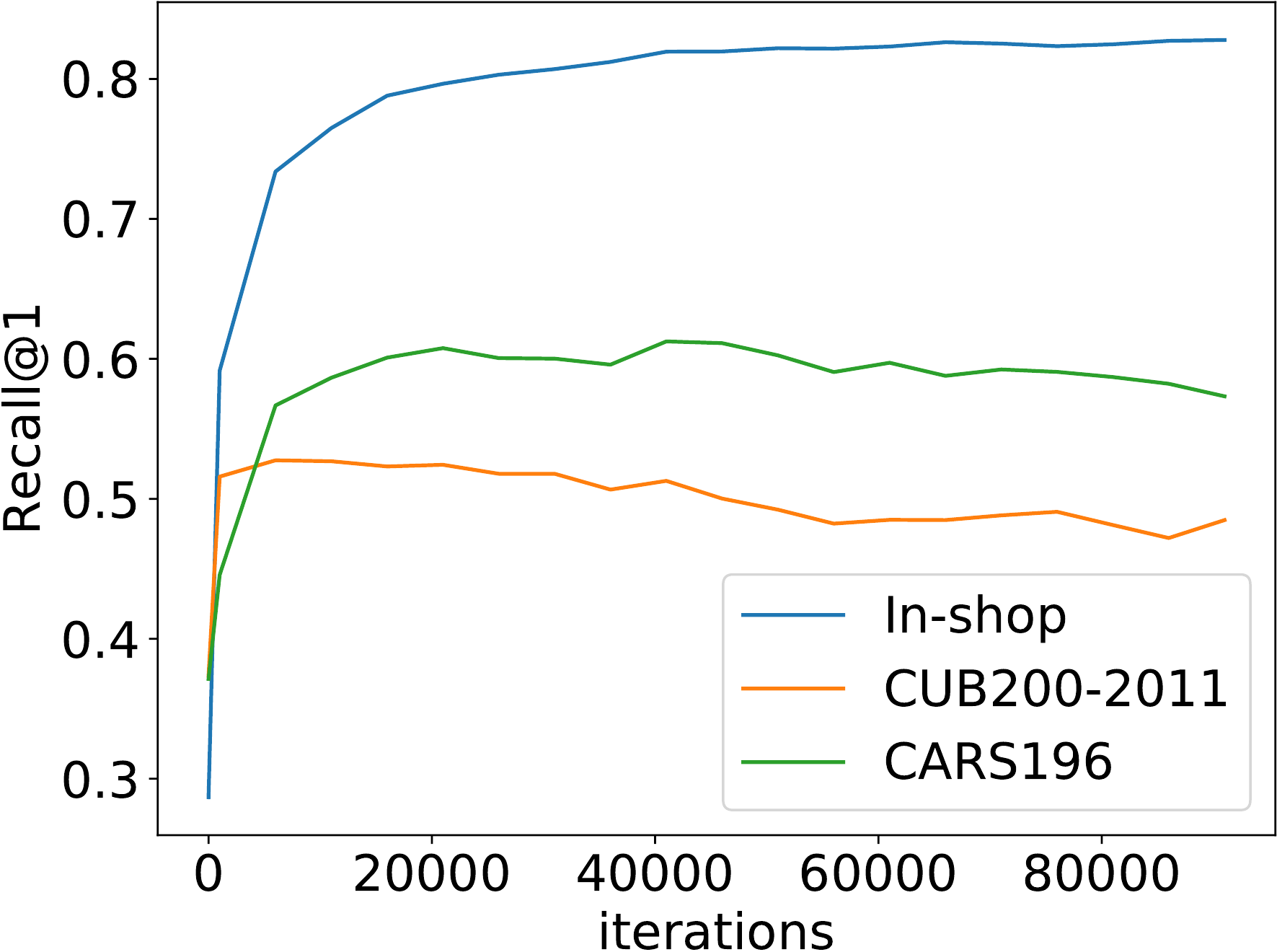}
}
\subfigure{\label{fig:balance}
  \includegraphics[width=0.28\columnwidth]{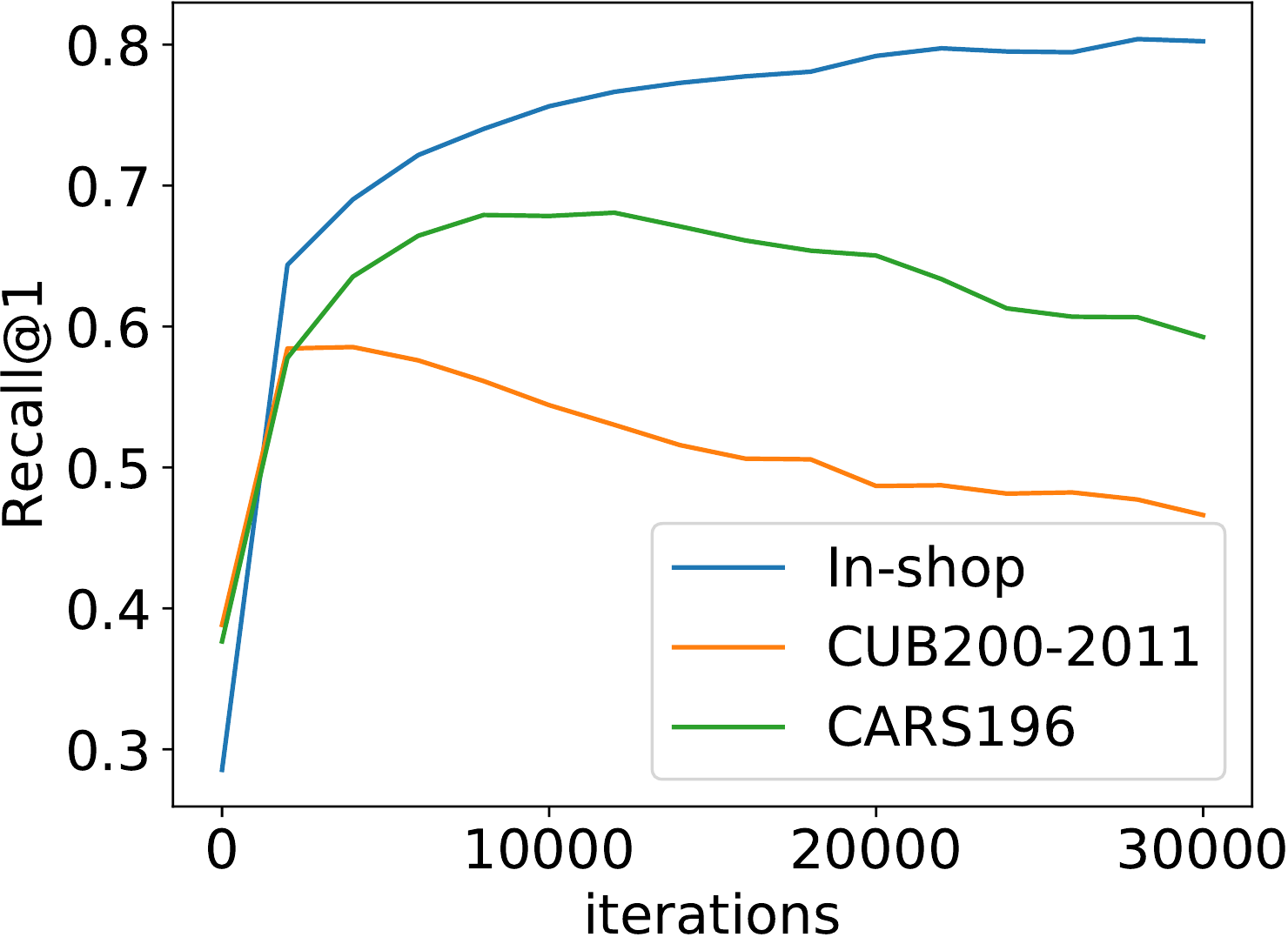}
}
\subfigure{\label{fig:kde}
  \includegraphics[width=0.28\columnwidth]{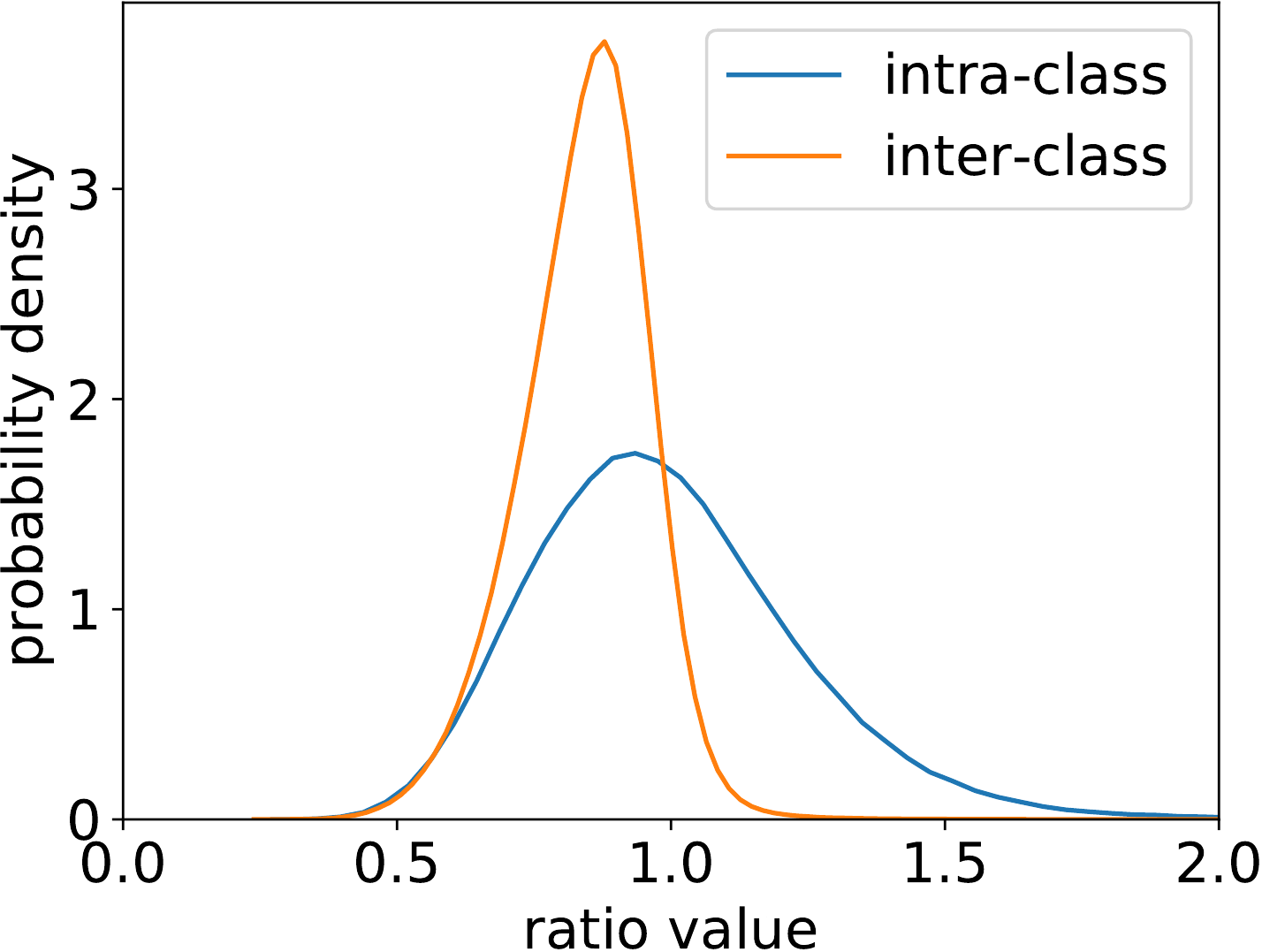}
}
  \setlength{\abovecaptionskip}{-0.1cm} 
  \caption{(a) We fuse the training images from the  CUB200-2011~\cite{wah2011caltech}, CARS196~\cite{KrauseStarkDengFei-Fei_3DRR2013}, and In-shop~\cite{liuLQWTcvpr16DeepFashion} datasets and train an embedding model using triplet loss \cite{schroff2015facenet}. The Recall@1 values computed over the evaluation set of each dataset are shown. We can see that the Recall@1 of CUB200-2011 and CARS196 soon begin to decrease, meaning overfitting is happening. However, the Recall@1 of In-shop keeps increasing, making it impossible to choose a stopping point having good performances on all datasets. (b) \yf{Same with (a) except that the BAL (Sec.~\ref{sec:exp}) sampling method is used.} (c) The distribution of the ratio between the distances measured by two models, \textit{i.e.} ImageNet specialist, and CUB200-2011 specialist, are visualized. The distances are computed over the CUB200-2011 images. We can see that the inter-class distance shrinks (ratio~$<1$) and the distances computed by the two models are not proportional (the ratio is not a constant).
  }
  \label{fig:2}
\vspace{-0.7cm}
\end{figure}

In this paper, we aim to train a universal embedding model to provide good performance on multiple domains. The universal embedding model should match the performance of multiple specialists on each of the specialist's domain. It is nontrivial to achieve this goal. Simply fusing the training data from different domains and training using existing methods will not obtain performances as good as the specialists. Since some domains are getting overfitted much faster than other domains, it is impossible to choose a stopping point where the universal embedding gets the best performance for all the domains, as illustrated in Fig.~\ref{fig:overfitting} and Fig.~\ref{fig:balance}. Another problem is that it becomes more difficult to sample effective training pairs or triplets after fusing multiple domains. The images across different domains  generally have larger differences than within the same domain. For the triplet loss~\cite{weinberger2006distance}, it will be too easy for a training triplet to produce no gradients if the negative image and anchor image are from different domains. These ineffective triplets will form the majority of all the triplets when fusing multiple domains, which makes it more difficult to see effective triplets during training. \yf{One may suspect that data imbalance may also be a reason for the performance decrease in Fig.~\ref{fig:overfitting}, but we do not believe it is the case because of two facts: 1) the proposed method in this paper also suffers from data imbalance but still achieves good performance; and 2) a data-balanced baseline we add in the experiment also has the early overfitting issue, as shown in Fig.~\ref{fig:balance}.}

To solve the early overfitting issue, we propose to distill the knowledge learned in specialists into a universal embedding. By distilling the knowledge from properly trained specialists, the obtained universal embedding will not overfit on any domain. Existing embedding knowledge distillation methods~\cite{liu2019knowledge,park2019relational,yu2019learning} are based on an assumption that the distances between pairs of images measured by a teacher model are exactly the same or proportional to the distances measured by a student model. %
This assumption does not hold for some cases when distilling the knowledge from specialists into a universal embedding. For example, we may want to unify a specialist trained on CUB200-2011~\cite{wah2011caltech} and another specialist trained on ImageNet~\cite{deng2009imagenet}, as shown in Fig.~\ref{fig:1}. We hope that the resulted universal embedding is roughly the same as the ImageNet specialist for non-bird sub-space. But for the bird sub-space, we expect the universal embedding can learn how to distinguish fine-grained birds from the CUB200-2011 specialist. The bird images only take a small part of the embedding space of the ImageNet specialist, but the bird images take a larger embedding space of the bird specialist. Due to the different bird space sizes, the pairwise distances between bird images measured by the CUB200-2011 specialist need to shrink so that the distances can fit in the ImageNet specialist. Fig.~\ref{fig:kde} shows the kernel density estimation of the ratio between the bird image distances measured by the ImageNet specialist and CUB200-2011 specialist.

To distill distances with the necessary shrinkage, we employ stochastic neighbor embedding (SNE)~\cite{hinton2003stochastic}. In SNE, the absolute pairwise distances are transformed into distributions and the Kullback-Leibler divergence~\cite{kullback1951information} is used to match the distributions of high-dimension embedding and low-dimension embedding. By doing so, the distance shrinkage is properly handled. 

In summary, our contributions are four-fold:
\setlist{nolistsep}
\begin{itemize}[noitemsep]
	\item We identify an important but unexplored embedding task: how to train a universal image embedding model to match the performance of several specialists on each of the specialist's domain.
	\item We propose to use distillation to avoid the early overfitting of some domains when training the universal embedding model.
	\item We distill the knowledge in embedding models using SNE, which properly handles distance shrinkage.
	\item We validate the effectiveness of the proposed universal embedding method in experiments using different combinations of several public datasets.
\end{itemize}

\section{Related Work}
\textbf{Deep Metric Learning} Recent deep metric learning research mainly falls into three directions: loss function, sampling policy, and learning with proxy. 
Contrastive loss~\cite{chopra2005learning} aims to pull similar sample pairs closer to each other and push dissimilar sample pairs farther away. However, directly minimizing the absolute distances of similar pairs to zero may be too restrictive. Triplet loss~\cite{weinberger2006distance} was proposed to solve this issue. In triplet loss, the relative distance order between anchor-positive and anchor-negative is ensured. %
Schroff \etal~\cite{schroff2015facenet} proposed semi-hard sampling to find the first negative farther than the positive to the anchor. Wu \etal~\cite{wu2017sampling} proposed to sample negative samples reciprocal to the anchor-negative distance. They showed that their sampling policy leads to a more balanced distribution of anchor-negative distance. %
Instead of sampling positives and negatives from a large pool of candidates, Movshovitz \etal~\cite{movshovitz2017no} proposed to use a proxy to represent a class of samples. The training speed and model performance were both improved by introducing proxies.

Existing research on deep metric learning mainly focuses on training a specialist embedding model for one domain of images. We are interested in training a universal embedding model having good performance on multiple domains.

\textbf{Knowledge Distillation} Initially, Bucilua \etal~\cite{bucilua2006model} compressed large ensemble models into smaller and faster models. The ensemble was used to label a large unlabeled dataset. Thereafter, the small neural network was trained using the ensemble labeled data. Hinton \etal~\cite{hinton2015distilling} improved the compression method by introducing a temperature to reduce the effect of large negative logits. \yf{Passalis and Tefas~\cite{passalis2018learning} turned the learned knowledge into a probability distribution and then transferred the probability distribution. The formulation of PKT~\cite{passalis2018learning} is very similar to the proposed method in this paper. The main differences between~\cite{passalis2018learning} and this paper are that we are solving a different task and handling the embedding distance shrinkage problem during distillation.}

Recently, several methods~\cite{liu2019knowledge,park2019relational,yu2019learning} tried to apply distillation to image embedding. They first trained a large teacher embedding model using existing methods and then distilled the knowledge learned by the teacher to a small student model. Instead of distilling the learned embedding vectors, the learned distances between embedding vectors are distilled. \yf{Different from~\cite{liu2019knowledge,park2019relational,yu2019learning}, we aim to train a universal embedding in this paper.} SNE~\cite{hinton2003stochastic} is used for embedding knowledge distillation, which can distill learned distances with shrinkage.

\textbf{Unifying Models} Gao \etal trained a classification model with an extremely large number of classes~\cite{gao2017knowledge}. This was done by distilling the knowledge from a hierarchy of smaller models. Vongkulbhisal \etal~\cite{vongkulbhisal2019unifying} developed a new task called Unifying Heterogeneous Classifiers (UHC) because of privacy considerations. %
The task to be solved in this paper is similar to UHC in that both tasks aim to unify several models into one. This paper unifies image embedding models, while UHC unifies classification models.

\section{Embedding Distillation}
We learn the universal embedding by distilling the knowledge from the specialist embeddings of different domains. In this section, we introduce the distillation techniques used in the paper. We first briefly review relational knowledge distillation (RKD)~\cite{park2019relational,yu2019learning} and then introduce the knowledge distillation method based on SNE~\cite{hinton2003stochastic}.

\subsection{Relational Knowledge Distillation}
In RKD, there is a teacher model and a relatively smaller student model, both of which are deep neural networks typically. %
The teacher model has better performance than the student model trained on the same training data without distillation. During distillation, the student model is trained to mimic the teacher model so that the performance of the student model could be similar to that of the teacher model. 
As a result, the student model obtained by distillation will be better than another student model trained without distillation. 

Let $\bx_i$, $\bx_j$, and $\bx_k$ denote three different images while $f_t(\cdot)$ and $f_s(\cdot)$ represent the teacher model and the student model, respectively. For simplicity, we also define the embedding vectors for $\bx$ computed by the teacher model and student model as $\bt = f_t(\bx)$ and $\bs = f_s(\bx)$, respectively. 
The relation distillation objective is to let the student mimic the learned distance between two embedding vectors from the teacher, which is given by
\begin{equation}
    \mathcal{L}_{RKD} = l_\delta(d_t,d_s),
\end{equation}
where $l_\delta$ could be $\ell$1-distance loss or Huber loss~\cite{huber1992robust}. $d_t=\frac{1}{\mu_t}\|\bt_i-\bt_j\|_2$ and $d_s=\frac{1}{\mu_s}\|\bs_i-\bs_j\|_2$ are the pairwise distances between two images measured by the teacher model and student model, respectively. $\mu_t$ and $\mu_s$ are the mean distance between images in a batch. A basic assumption behind this mean distance normalization is that $\|\bt_i-\bt_j\|_2$ should be proportional to $\|\bs_i-\bs_j\|_2$. %
For angle-wise distillation~\cite{park2019relational}, similar triangles are constructed, which is also based on the proportional assumption. In summary, existing relational knowledge distillation methods are designed for cases where the absolute distances between a pair of images measured by the teacher model and student model are the same or proportional.

\begin{figure}[t]
\vspace{-1mm}
  \centering
  \includegraphics[width=0.5\columnwidth]{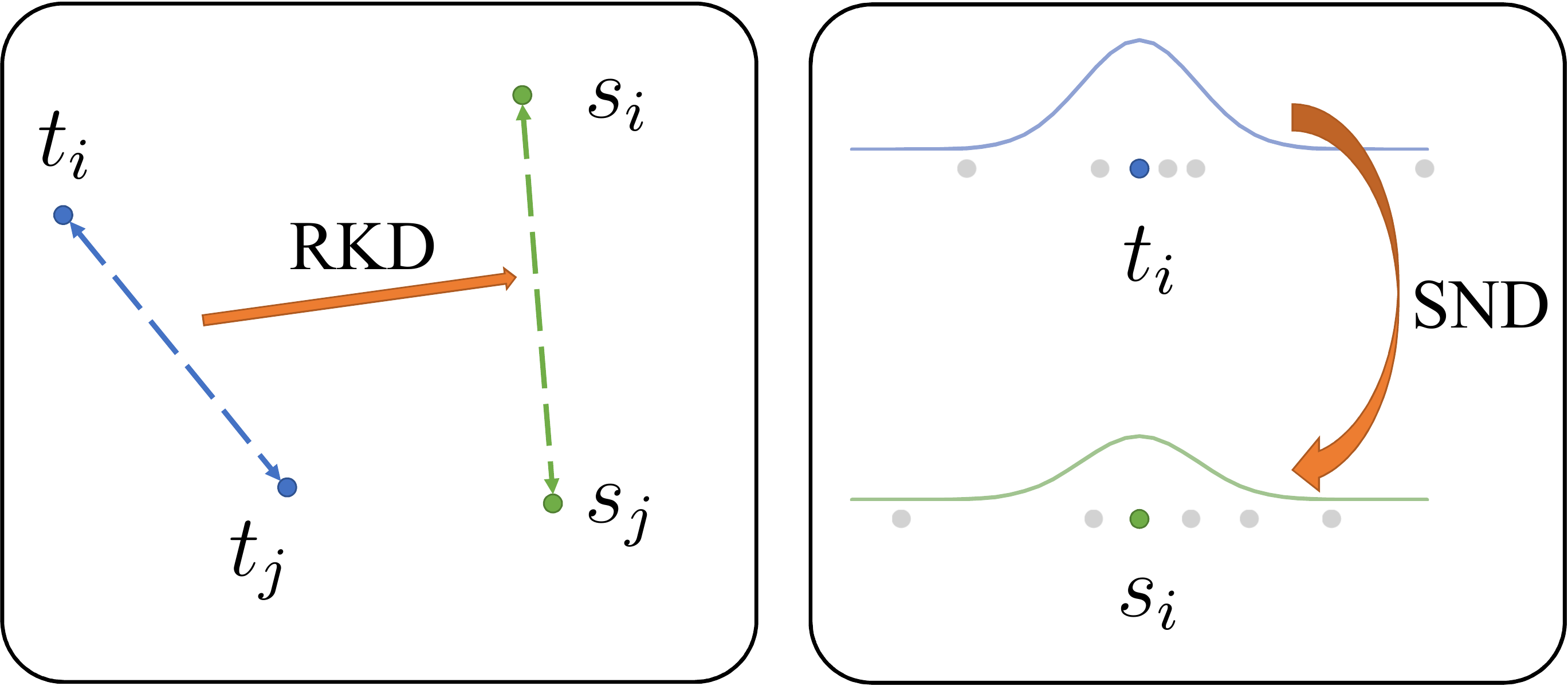}
  \setlength{\abovecaptionskip}{-0.05cm} 
  \caption{Relational knowledge distillation~(RKD) and stochastic neighbor distillation~(SND). Instead of directly distilling the absolute distances between images, the distance probability distribution is distilled in SND.}
  \label{fig:snd}
\vspace{-0.5cm}
\end{figure}

\subsection{Stochastic Neighbor Distillation}
As stated in the introduction, when unifying a fine-grained dataset, \eg CUB200-2011, and a large dataset, \eg ImageNet, the embedding sub-space taken by the CUB200-2011 images will shrink compared with the embedding sub-space taken by CUB200-2011 in a CUB200-2011 specialist. The shrinkage is non-uniform, as the distance ratio is not a constant. This breaks the proportional assumption in RKD.

To address this issue, inspired by SNE~\cite{hinton2003stochastic} and PKT~\cite{passalis2018learning}, we distill knowledge by a distance distribution instead of absolute distance. The proposed method is named stochastic neighbor distillation (SND), which is shown in Fig.~\ref{fig:snd}.

We use the SNE objective to train a new embedding model instead of just computing vectors for certain objects. The high-dimension and low-dimension vectors in~\cite{hinton2003stochastic} can be viewed as the output of the teacher and student models, respectively. Considering the image $\bx_i$ with feature $\bt_i$, the probability that it would pick $\bx_j$ with feature $\bt_j$ as its neighbor measured by the teacher model is
\begin{equation}
\label{eq:p}
    p_{ij} = \frac{\exp(-\frac{\|\bt_i-\bt_j\|_2^2}{2\sigma_i^2})}
    {\sum_{k\neq i}\exp(-\frac{\|\bt_i-\bt_k\|_2^2}{2\sigma_i^2})},
\end{equation}
where $\sigma_i$ is the variance of the Gaussian kernel used in the $i$-th distance distribution for the teacher embedding, denoted by $P_i$. 
We can set the value of $\sigma_i$ by hand or use binary search to determine the value of $\sigma_i$ so that the perplexity of the distance distribution equals to a predefined value~\cite{hinton2003stochastic}.
Similarly, we can define the distribution for $\bx_i$ with feature $\bs_i$ picking $\bx_j$ with feature $\bs_j$ as its neighbor measured by the student model as:
\begin{equation}
\label{eq:q}
    q_{ij} = \frac{\exp(-\frac{\|\bs_i-\bs_j\|_2^2}{2{\sigma'}_i^2})}
    {\sum_{k\neq i}\exp(-\frac{\|\bs_i-\bs_k\|_2^2}{2{\sigma'}_i^2})},
\end{equation}
where $\sigma'_i$ is the variance of the Gaussian kernel used in the $i$-th distance distribution for the student embedding, denoted by $Q_i$. The distillation objective is given by
\begin{equation}
     \mathcal{L}_{SND} = \sum_i\sum_jp_{ij}log\frac{p_{ij}}{q_{ij}}
     =\sum_iKL(P_i \parallel Q_i).
\end{equation}
To better understand how this loss behaves, the gradient of the loss with respect to $\bs_i$ is given in~\cite{hinton2003stochastic}:
\begin{equation}
    \frac{\partial \mathcal{L}_{SND}}{\partial \bs_i} = 2 \sum_j (\bs_i - \bs_j)(p_{ij} - q_{ij} + p_{ji} - q_{ji}).
\end{equation}
From the gradient, we can see that $\bs_i$ is either pulled towards $\bs_j$ or pushed away from $\bs_j$ depending on whether $\bx_j$ is observed to be a neighbor of $\bx_i$ more or less often than expected. SND inherits two nice properties from SNE: 1) The absolute distances between images are turned into probabilities, which properly handles the distance shrinkage; and 2) By tuning the variance of the Gaussian, it is possible to control how many neighbors that are considered during the distillation. If small variance values are used, SND will preserve the local structure of the data manifold. When the variances are large enough, SNE is equivalent to minimizing the mismatch between squared distances in the two spaces \cite{hinton2003stochastic}. Therefore, SND can be viewed as a generalization of RKD.

\section{Universal Embedding Training}
\label{sec:uni}
There are two cases to consider: 1) unifying some mutually exclusive domains; and 2) unifying a coarse-grained domain with its fine-grained sub-domains, \eg ImageNet and CUB200-2011. 
A naive way to train the universal embedding is to fuse the training images from multiple domains together. The label correspondence across domains also needs to be figured out. Fusing and training with single domain methods cannot obtain a good universal embedding, because some domains are getting overfitted much sooner than other domains, as shown in the introduction. 

We propose to distill the knowledge from properly trained specialist embedding models to the universal embedding. Let $D_i|_{i=1}^m$ denote the domains that need to be unified, where $m$ is the number of domains. For each domain $D_i$, we first train a specialist embedding model $f_t^i$ using existing single domain methods. All the specialist embedding models are teachers to the universal embedding. In SNE, the distance distribution is estimated over the whole dataset, which is computational infeasible for large datasets like ImageNet. Therefore, we sample a mini-batch and use all the distances inside a mini-batch to approximate the distance distribution.

As mini-batch is used, we must design a sampling policy for $m$ domains of data. Each specialist embedding model is trained with images in only one domain, so the specialist can only be used to encode the images in that domain. As a result, we are unable to compute the distances between images across domains. Based on this fact, we have each mini-batch only containing images from one domain, which is named \textit{domain-specific sampling}. The frequency of choosing one domain to form a mini-batch is proportional to the number of images in that domain. 
After determining which domain to use, we follow the convention to choose the images inside a mini-batch. We randomly select $c$ classes and sample $k$ images for each class to form a mini-batch with size $c\times k$, where $k$ is set to $4$ in this paper. With the domain-specific mini-batch sampling, we minimize the $\mathcal{L}_{SND}$ between one specialist and the universal embedding in each training iteration.

\yf{Without explicitly informing the universal embedding model that images from different domains are dissimilar, the universal embedding model may mix different domains together. In the experiment, we show that the proposed universal embedding model does not mix different domains.} One possible reason may be that we are using a CNN which is pre-trained for ImageNet classification. The ResNet~\cite{he2016deep} features of images from different domains are already well separated. During the embedding training process, the objective is to make similar images have closer embeddings. As a result, the trained deep embedding model has the property of mapping similar images to similar sub-spaces. This property can act as a regularization to avoid mapping dissimilar images to the same embedding.

\section{Experiments}
\label{sec:exp}
We evaluate the proposed method by training a universal embedding model for multiple domains. Each one of ImageNet~\cite{deng2009imagenet}, CUB200-2011~\cite{wah2011caltech}, CARS196~\cite{KrauseStarkDengFei-Fei_3DRR2013}, In-shop Clothes~\cite{liuLQWTcvpr16DeepFashion}, Stanford Online Produces~\cite{oh2016deep}, and PKU VehicleID~\cite{liu2016deep} can be viewed as a domain. We first describe the experimental settings. Then, we show that the universal embedding does not mix different domains. In the end, the unification results are given.

\textbf{ImageNet} is used to refer to ILSVRC2012~\cite{deng2009imagenet}. It is widely used for object recognition. In this paper, we use ImageNet for the image embedding task for its variety of categories. The $1,281,167$ training images are used for training and the $50,000$ validation images are used for testing.

\textbf{CUB200-2011} contains $11,788$ images belonging to $200$ bird categories. The first $100$ categories ($5,864$ images) are used for training and the remaining $100$ categories ($5,924$ images) are used for testing. The bounding box annotations are not used for training or testing.

\textbf{CARS196} consists of $16,185$ images of $196$ classes of cars. The first $98$ categories ($8,054$ images) are used for training and the remaining $98$ categories ($8,131$ images) are used for testing. The bounding box annotations are not used for training or testing.

\textbf{In-shop Clothes Retrieval (In-shop)} contains cloth images with large pose and scale variations. We follow the settings used in~\cite{liuLQWTcvpr16DeepFashion}. $3,997$ categories ($25,882$ images) are used for training and $3,985$ categories ($26,830$) are used for testing. The testing set is further divided into a gallery set ($12, 612$ images) and probe set ($14,218$ images).

\textbf{Stanford Online Products (SOP)} has $120,053$ images of $22,634$ classes of online products. We follow the settings used in~\cite{oh2016deep}, where $11,318$ categories ($59,551$ images) are used for training and the remaining $11,316$ categories ($60,502$ images) are used for testing. 

\textbf{PKU VehicleID (VID)} is a large-scale dataset for vehicle re-identification. It contains $221,736$ images of $26,267$ vehicles captured by surveillance cameras. We follow the settings used in~\cite{liu2016deep}. $13,134$ vehicles are used for training. There are three test sets of different sizes defined in~\cite{liu2016deep}. We only use the large test set containing $20,038$ images of $2,400$ vehicles.

In all the experiments, we use ResNet50 pre-trained for ImageNet classification task as the backbone CNN. 
The training images are resized to $256\times 256$ and then randomly cropped to $224\times 224$. Randomly horizontal flipping is also used for data augmentation. Central crop is used for the test images. We add a fully-connected layer to project the 2048-dim ResNet50 output into a 128-dim embedding vector and further normalize the embedding vector to unit-length. To train a specialist, we either use Triplet semi-hard~\cite{schroff2015facenet} or Multi-Similarity~\cite{wang2019multi}. The batch size is set to $128$ and Adam optimizer with learning rate $1e^{-5}$ is used for training. \yf{Since there is no validation set in the previous data split setting, we tune $\sigma$ by self-distillation on one dataset and then use the obtained $\sigma$ value for the experiments on the other datasets.} The Recall@k are reported for performance comparison. The proposed method is implemented by Tensorflow~\cite{tensorflow2015-whitepaper}. Nvidia GTX 1080Ti GPUs are used to train the embedding models.

\yf{We design four baseline methods for comparison. In the ``Concatenation'' baseline, we concatenate the embedding vectors extracted by the trained specialists and then use PCA to project the concatenated embedding into 128-dim. The remaining three baselines can be summarized as fusing the training data and then training with single domain methods. The difference between these three baselines lies in the sampling method: 1) na\"ive sampling; 2) domain-specific (DS) sampling introduced in Sec.~\ref{sec:uni}; and 3) domain-balanced (BAL) sampling. In na\"ive sampling, the training images from different datasets are mixed in each training batch. BAL sampling is based on DS sampling, but the datasets are sampled with equal probability regardless of the number of training images from each dataset.}

\subsection{Visualization of the Embedding}
\label{sec:no_confusion}
In this section, \yf{we show that the proposed universal embedding model does not mix different domains together, even though domain-specific sampling is used. Instead, it preserves the separation of different domains while matching the performance of several specialists on each specialist’s domain with a single embedding model.} We first visualize both the specialist embedding and the universal embedding with t-SNE and then show the principal component analysis results of the embedding vectors.

\begin{figure}[t]
\vspace{-2mm}
  \centering
\subfigure{
  \includegraphics[width=0.24\columnwidth]{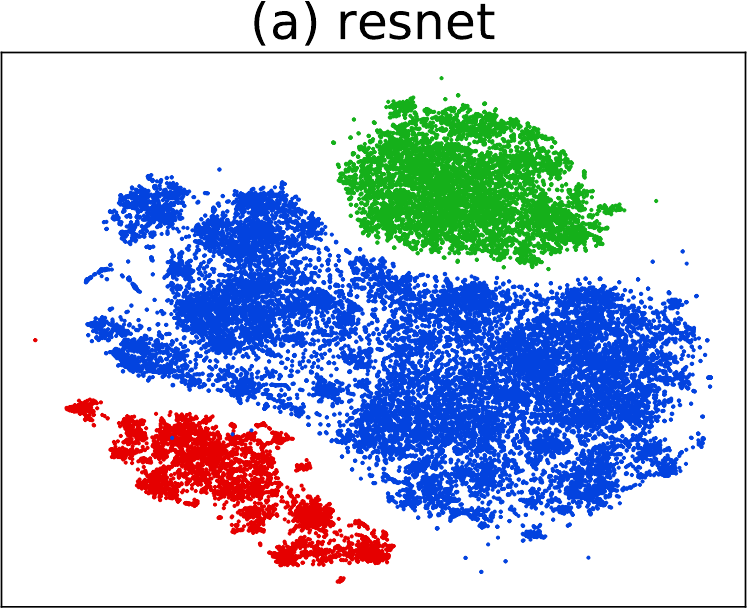}
  \includegraphics[width=0.24\columnwidth]{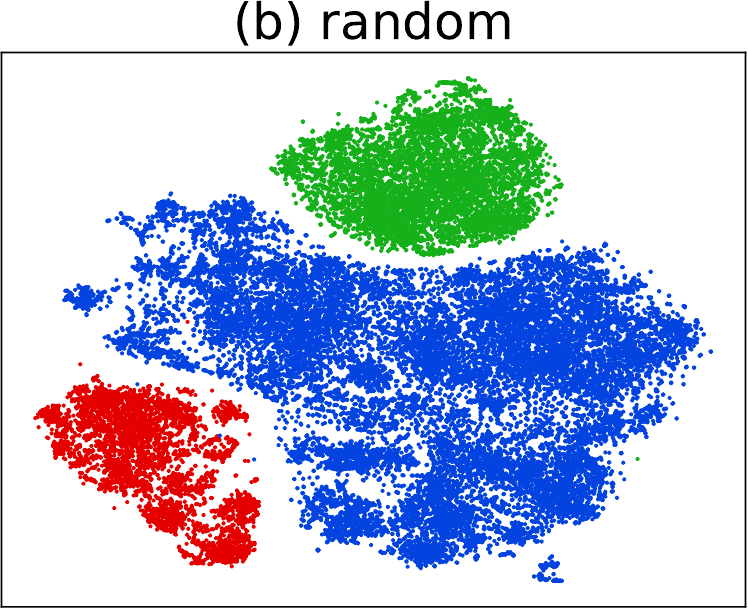}
  \includegraphics[width=0.24\columnwidth]{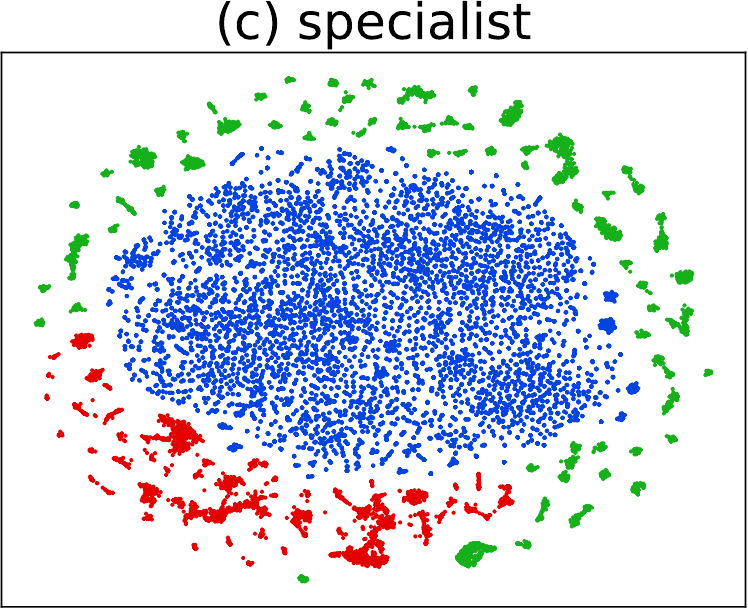}
  \includegraphics[width=0.24\columnwidth]{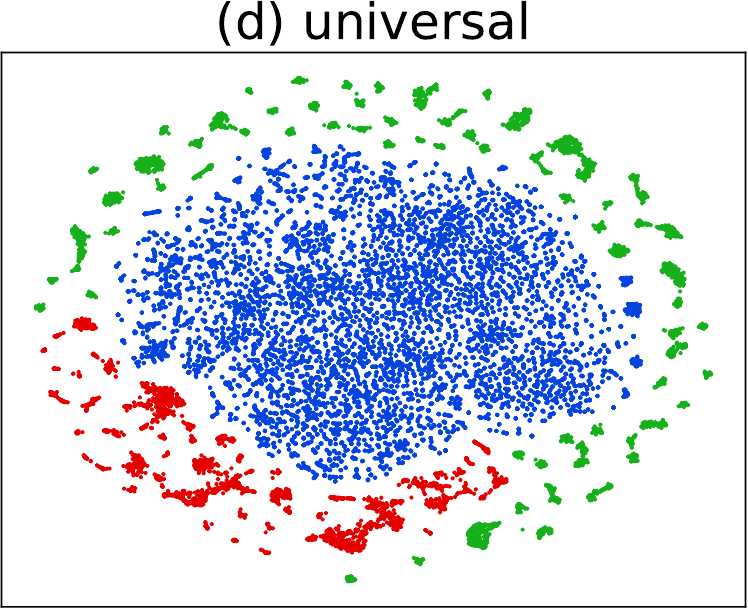}
}
  \setlength{\abovecaptionskip}{-0.4cm} 
  \caption{t-SNE visualization of CUB200-2011 (red), CARS196 (green), and In-shop (blue) embeddings. The embeddings in different figures are computed by different models: (a) ResNet50 pre-trained on ImageNet; (b) ResNet50 pre-trained on ImageNet followed by a randomly initialized 128-way fully-connected layer; (c) the specialist embedding model corresponding to each domain; (d) a universal embedding model.}
  \label{fig:tsne}
\vspace{-0.3cm}
\end{figure}

\begin{figure}[t]
  \centering
\subfigure{
  \includegraphics[width=0.28\columnwidth]{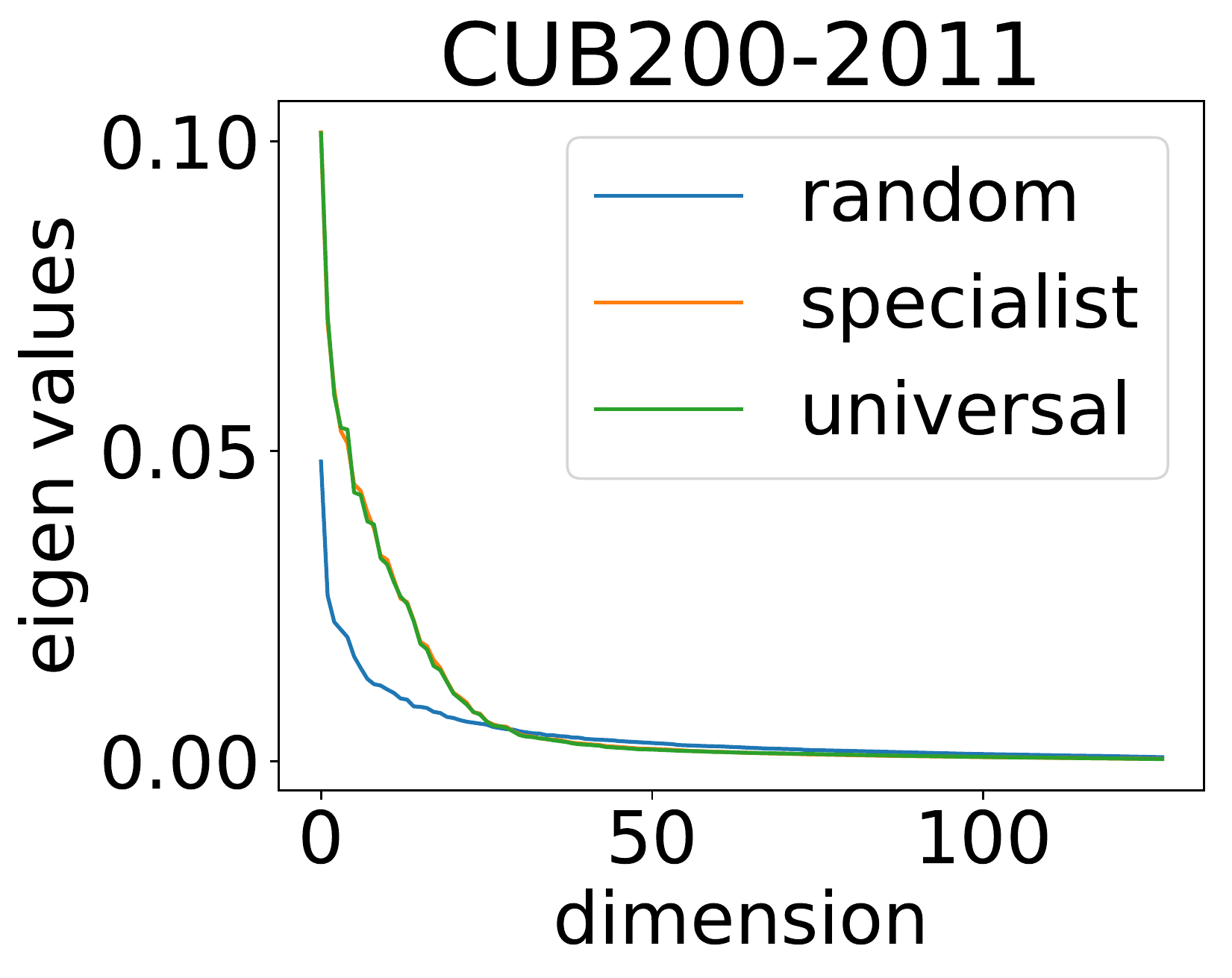}
  \includegraphics[width=0.28\columnwidth]{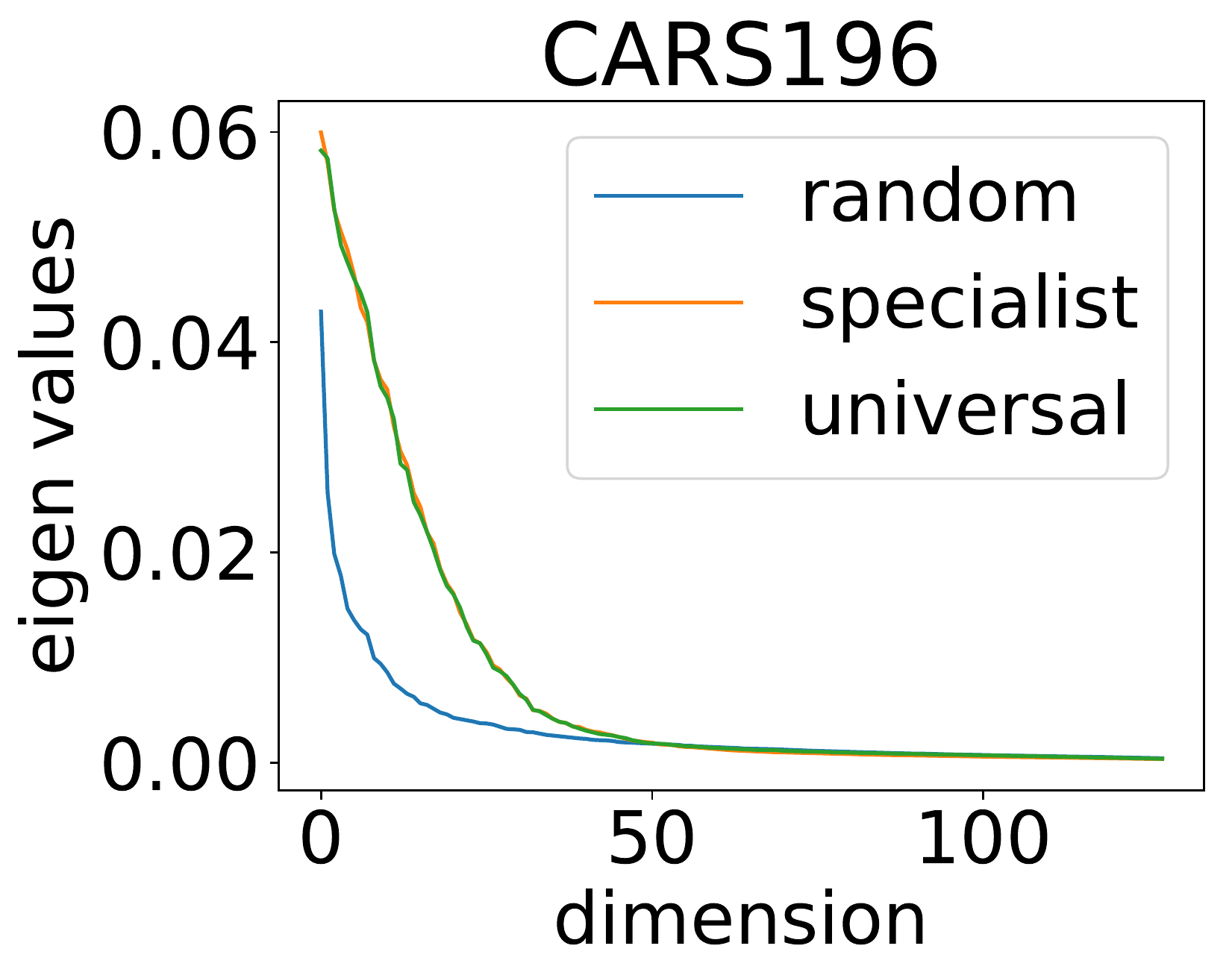}
  \includegraphics[width=0.28\columnwidth]{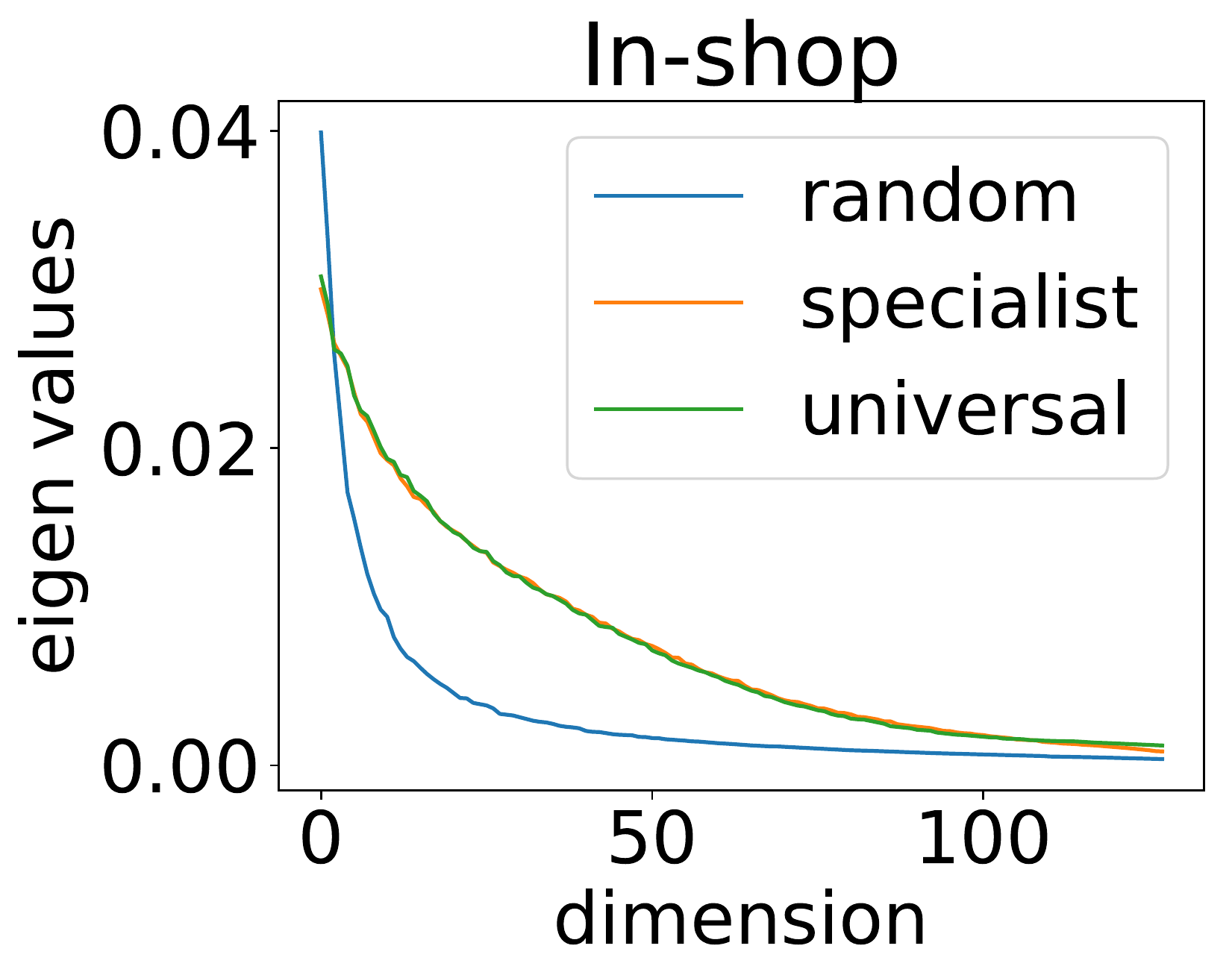}
}
  \setlength{\abovecaptionskip}{-0.2cm} 
  \caption{Eigen values in the PCA of the embedding vectors computed by the three models for three datasets. We have two observations: 1) the embedding vectors lie in a low-dimension sub-space; 2) the eigen values are almost the same for the specialists and the universal embedding.}
  \label{fig:pca}
\vspace{-0.5cm}
\end{figure}

\textbf{t-SNE Visualization} In Fig.~\ref{fig:tsne}, we show four t-SNE visualizations of the CUB200-2011, CARS196, and In-shop. %
The perplexity is set to $50$ when generating the figure. In Fig.~\ref{fig:tsne}(a), we visualize the embedding vectors generated by a pre-trained ResNet50. We can see that the ResNet50 features of the images from different datasets are already well separated. In Fig.~\ref{fig:tsne}(b), we visualize the embedding vectors generated by a randomly initialized model. The weights in the fully-connected layer are drawn from a Gaussian distribution and weights in ResNet50 layers are pre-trained by ImageNet. It can be observed that the fully-connected layer preserves the relationships of the images. %
In Fig.~\ref{fig:tsne}(c), we train a specialist embedding for each dataset and use the corresponding specialist to extract image embeddings. \yf{The three specialists are trained independently, meaning that each specialist is not aware of the other two.} We put all the image embeddings from the three datasets together and compute the t-SNE visualization. %
There are still clear boundaries between the datasets. The specialists are projecting images from different domains into different sub-spaces. In Fig.~\ref{fig:tsne}(d), we visualize the embeddings computed by a universal embedding trained with the proposed method. It can be observed that the images from the three datasets are mapped to different sub-spaces by the universal embedding. %
From these results, we infer that each dataset only takes certain sub-space in the embedding space and the whole embedding space is large enough to place all the three datasets in different sub-spaces.

In some cases of unifying mutually exclusive domains, the embeddings learned by different specialists may not be as well separated as in Fig.~\ref{fig:tsne}(c). Further study is needed to know how good distillation will work for such cases. We leave that to future work. When unifying a coarse-grained domain (ImageNet) with a fine-grained sub-domain (CUB200-2011), the labels in the coarse-grained domain can guide the universal model to separate the sub-domain (bird) images from other (non-bird) images. 

\textbf{Principal component analysis} We also visualize the eigen values in the PCA of the embeddings of these three datasets computed by three models, \textit{i.e.} the random model, the specialist, and the universal embedding. From Fig.~\ref{fig:pca}, we can see that most of the eigen values in the left two figures are close to zeros, meaning that CUB200-2011 and CARS196 datasets lie in low-dimension sub-spaces. This may be because there are many bird categories and car categories in ImageNet. After pre-training, ResNet50 maps bird images and car images into a low-dimension sub-space. For In-shop, there are more eigen values much larger than zeros, which may be because there are not many cloth related categories in ImageNet. The second observation from Fig.~\ref{fig:pca} is that the eigen values of embeddings computed by the specialist and the universal embedding are almost the same. This is in accordance with the conclusion from the t-SNE visualization that the whole embedding space is large enough to place all the three datasets in different sub-spaces without shrinking anything.

\subsection{Unifying Mutually Exclusive Domains}
\label{sec:exclusive}
In this section, we train two universal models, one by unifying the CUB200-2011, CARS196, and In-shop specialists, and the other by unifying the In-shop, SOP, and VID specialists.

\begin{table}[t]
  \setlength{\belowcaptionskip}{1pt}
  \caption{Performance comparisons of specialists and universal embedding models. The recalls are computed without fusing the evaluation sets.}
  \label{tab:in_domain}
  \centering
  \begin{tabular}{l|cc|cc|cc||cc|cc|cc}
    \toprule
    & \multicolumn{6}{c||}{Combination 1} & \multicolumn{6}{c}{Combination 2} \\
     & \multicolumn{2}{c|}{CUB200} & \multicolumn{2}{c|}{CARS196} & \multicolumn{2}{c||}{In-shop} & \multicolumn{2}{c|}{SOP} & \multicolumn{2}{c|}{VID} & \multicolumn{2}{c}{In-shop}\\
      & R1 & R2 & R1 & R2 & R1 & R2 & R1 & R2 & R1 & R2 & R1 & R2 \\
    \midrule
    Specialists & 59.8 & 71.6 & 70.1 & 79.4 & 84.3 & 89.2 & 73.4 & 78.8 & \textbf{86.9} & \textbf{90.8} & 84.3 & 89.2\\
    Ours\_RKD & \textbf{61.0} & \textbf{72.1} & \textbf{73.3} & \textbf{82.2} & \textbf{85.2} & \textbf{90.2} & \textbf{74.2} & \textbf{79.2} & 86.8 & 90.6 & \textbf{84.6} & \textbf{89.7}\\
    Ours\_SND & 59.7 & 72.0 & 71.8 & 81.2 & 84.3 & 89.5 & 73.7 & 78.8 & \textbf{86.9} & 90.5 & 82.9 & 88.3 \\ 
    \bottomrule
  \end{tabular}
\vspace{-0.5cm}
\end{table}

\textbf{Evaluation on single domain} We first report the performance of the specialist embedding models obtained by Triplet semi-hard and the performance of the universal embeddings on each dataset separately. The results are shown in Table~\ref{tab:in_domain}. 
The specialist row shows the performance of the specialist embedding models in each of their specific domain respectively. ``Our\_RKD'' and ``Ours\_SND'' rows show the performances of the universal embedding models trained with RKD and SND by distilling three different specialist models. Note that the specialist row shows the performances of \emph{three different specialist} models in their specific domain, while the ``Our\_RKD'' and ``Ours\_SND'' rows show the performance of one universal embedding in three different domains. \yf{To achieve the performance in the specialist row in a real image search scenario, an additional step of determining which specialist to use for a user uploaded query image is needed.}

That ``Ours\_RKD'' achieves better performance than ``Ours\_SND'' may be due to two reasons. First, as shown in Sec.~\ref{sec:no_confusion}, the whole embedding space is large enough to place all the three datasets in different areas without any shrinkage. Second, we are using fixed $\sigma$ values instead of binary search. It is interesting to find that the performance of ``Ours\_RKD'' is even slightly better than the specialists. \yf{This may be because the unifying process is similar to three independent self-distillations~\cite{furlanello2018born}}, which can produce a better student model than the teacher model~\cite{park2019relational}. %

\begin{table}[t]
\small
  \setlength{\belowcaptionskip}{1pt}
  \caption{Performance comparisons of unifying CUB200-2011, CARS196, and In-shop datasets. The recalls are computed over the fused evaluation set.}
  \label{tab:cci}
  \centering
  \begin{tabular}{c|l|ccc|ccc|ccc|ccc}
    \toprule
     & & \multicolumn{3}{c|}{CUB200-2011} & \multicolumn{3}{c|}{CARS196} & \multicolumn{3}{c|}{In-shop} & \multicolumn{3}{c}{Average} \\
     &  & R1 & R2 & R4 & R1 & R2 & R4 & R1 & R2 & R4 & R1 & R2 & R4 \\
    \midrule
    \parbox[t]{2mm}{\multirow{4}{*}{\rotatebox[origin=c]{90}{specialist}}} 
      & CUB200-2011 & 59.8 & 71.5 & 81.6 & 31.7 & 43.1 & 55.1 & 24.6 & 31.1 & 37.7 & 38.7  & 48.6 & 58.1 \\
      & CARS196 & 26.1 & 37.2 & 49.2 & 70.0 & 79.3 & 86.1 & 18.6 & 23.9 & 29.5 & 38.2 & 46.8 & 54.9 \\
      & In-shop & 7.6 & 11.9 & 18.7 & 13.9 & 19.0 & 25.5  & 84.1 & 89.1 & 92.7 & 35.2 & 40.0 & 45.6 \\
      & 3-way Concat. & 47.3 & 60.5 & 73.0 & 64.2 & 76.9 & 86.1 & 69.6 & 78.2 & 84.8 & 60.4 & 71.9 & 81.3 \\
    \midrule
    \parbox[t]{2mm}{\multirow{5}{*}{\rotatebox[origin=c]{90}{universal}}} 
    & Triplet \cite{schroff2015facenet} & 48.5 & 60.4 & 71.1 & 57.3 & 69.1 & 78.2 & 82.8 & 88.8 & 92.8 & 62.9  & 72.8  & 80.7 \\
    & Triplet+DS & 54.8 & 66.5 & 77.9 & 66.1 & 76.2 & 84.4 & 76.4 & 86.1 & 91.0 & 65.8 & 76.3 & 84.4 \\
    & Triplet+BAL & 57.3 & 69.4 & 78.8 & 67.2 & 77.4 & 84.9 & 71.6 & 78.9 & 84.8 & 65.4 & 75.2 & 82.8 \\
    & Ours\_RKD & \textbf{61.0} & \textbf{72.1} & \textbf{82.0} & \textbf{73.3} & \textbf{82.2} & \textbf{88.7}  & \textbf{85.2} & \textbf{90.2} & \textbf{93.4} & \textbf{73.2} & \textbf{81.5} & \textbf{88.0} \\
    & Ours\_SND & 59.7 & 72.0 & 81.5 & 71.8 & 81.2 & 87.6  & 84.3 & 89.5 & 93.2 & 71.9 & 80.9 & 87.4 \\
    \bottomrule
  \end{tabular}
\vspace{-0.2cm}
\end{table}

\begin{table}[t]
\small
  \setlength{\belowcaptionskip}{1pt}
  \caption{Performance comparisons of unifying SOP~\cite{oh2016deep}, VID~\cite{liu2016deep}, and In-shop~\cite{liuLQWTcvpr16DeepFashion} datasets. The recalls are computed over the fused evaluation set.}
  \label{tab:isv}
  \centering
  \begin{tabular}{c|l|ccc|ccc|ccc|ccc}
    \toprule
     & & \multicolumn{3}{c|}{SOP} & \multicolumn{3}{c|}{VID} & \multicolumn{3}{c|}{In-shop} & \multicolumn{3}{c}{Average}\\
     &  & R1 & R2 & R4 & R1 & R2 & R4 & R1 & R2 & R4 & R1 & R2 & R4 \\
    \midrule
    \parbox[t]{2mm}{\multirow{4}{*}{\rotatebox[origin=c]{90}{specialist}}} 
      & SOP & 73.3 & 78.6 & 82.9 & 38.8 & 45.3 & 51.3 & 40.1 & 47.8 & 55.3 & 50.7 & 57.2 & 63.2\\
      & VID & 38.5 & 42.8 & 47.1 & 86.7 & \textbf{90.8} & \textbf{93.5} & 18.8 & 24.1 & 29.5 & 48.0 & 56.2 & 56.7 \\
      & In-shop & 53.8 & 58.8 & 63.2 & 31.9 & 38.0 & 44.2 & 83.8 & 88.7 & 92.2 & 56.5 & 61.8 & 66.5\\
      & 3-way Concat. & 71.9 & 77.1 & 81.3 & 76.1 & 82.1 & 86.9 & 74.9 & 82.0 & 87.6 & 74.3 & 80.4 & 85.3 \\
    \midrule
    \parbox[t]{2mm}{\multirow{5}{*}{\rotatebox[origin=c]{90}{universal}}} 
    & Triplet \cite{schroff2015facenet} & 71.0 & 76.5 & 81.1 & 84.2 & 88.8 & 92.2 & 79.5 & 85.8 & 90.2 & 78.2 & 83.7 & 87.8 \\
    & Triplet+DS & 72.7 & 77.9 & 82.4 & 84.3 & 88.8 & 92.4 & 81.2 & 87.3 & 91.7 & 79.4 & 84.7 & 88.8 \\
    & Triplet+BAL & 72.5 & 77.2 & 81.4 & 75.0 & 81.7 & 87.1 & 83.1 & 88.1 & 91.6 & 76.9 & 82.3 & 86.7 \\
    & Ours\_RKD & \textbf{74.2} & \textbf{79.2} & \textbf{83.3} & 86.8 & 90.6 & \textbf{93.5} & \textbf{84.5} & \textbf{89.7} & \textbf{93.1} & \textbf{81.9} & \textbf{86.5} & \textbf{90.1} \\
    & Ours\_SND & 73.7 & 78.8 & 83.1 & \textbf{86.9} & 90.5 & 93.4  & 82.9 & 88.3 & 92.3 & 81.2 & 85.9 & 89.6 \\
    \bottomrule
  \end{tabular}
\vspace{-0.6cm}
\end{table}

\textbf{Evaluation on fused domains} Instead of evaluating in each domain separately, we consider a harder problem by fusing the evaluation set of all the unified domains together to simulate the real-world image search system handling a diversity of domains. To be specific, when computing the recall for CUB200-2011, the CUB200-2011 evaluation set is used as the probe and all the evaluation images from the three domains~(CUB200-2011, CAR196, In-shop) are used as the gallery. The In-shop dataset evaluation is slightly different from the other datasets because the evaluation images are divided into probe and gallery. For In-shop dataset evaluation, we keep the probe set unchanged and fuse the gallery set with the evaluation images of the other two datasets.

The results of unifying CUB200-2011, CARS196 and In-shop specialists are shown in Table~\ref{tab:cci}, \yf{and the results of unifying In-shop, SOP, and VID specialists are shown in Table~\ref{tab:isv}. In both tables, we first report the performance of all the specialists under the fused evaluation setting and then provide the performance of embeddings trained with the four baseline methods.} Finally, we report the performance of training universal embedding using distillation methods with the specialists. 

First, we compare the performance before (Table~\ref{tab:in_domain}) and after (Table~\ref{tab:cci}) fusing the evaluation set. The Recall@1 of In-shop specialist drops from $84.3$ to $84.1$ when evaluated on fused datasets. The specialist image embedding models make some mistakes on cross-domain images, but the confusion is not serious. For ``Ours\_RKD'' and ``Ours\_SND'', the performance drop is less than $0.1$, meaning the universal embedding model is able to distinguish images from different domains. \yf{The ``Concatenation'' baseline achieves poor results because two-thirds of the embedding are out of domain and this has a side effect on the distance.} The performance of ``Triplet'' trained with the fused dataset is much worse than the specialists in their specific domains, which may be because of the early overfitting and ineffective triplet problem. \yf{Domain-specific sampling can solve the ineffective triplet problem, so ``Triplet+DS'' is much better than ``Triplet''. Comparing ``Triplet+DS'' and ``Triplet+BAL'', we find that the domains with fewer images, \textit{i.e.} CUB200-2011 and CARS196, show improved results, but the domain with the most images, \textit{i.e.} In-shop, shows worse results.} The results of distillation methods are comparable to or better than the specialists in their specific domains, showing the effectiveness of training universal embedding using distillation.

\begin{table}[t]
\small
  \setlength{\belowcaptionskip}{2pt}
  \caption{Performance comparisons of unifying CUB200-2011, CARS196, and In-shop specialists trained by Multi-Similarity~\cite{wang2019multi}. The recalls are computed over the fused evaluation set.}
  \label{tab:distill_ms}
  \centering
  \begin{adjustbox}{max width=\columnwidth}
  \begin{tabular}{c|l|ccc|ccc|ccc|ccc}
    \toprule
     & & \multicolumn{3}{c|}{CUB200-2011} & \multicolumn{3}{c|}{CARS196} & \multicolumn{3}{c|}{In-shop} & \multicolumn{3}{c}{Average} \\
     &  & R1 & R2 & R4 & R1 & R2 & R4 & R1 & R2 & R4 & R1 & R2 & R4 \\
    \midrule
    \parbox[t]{2mm}{\multirow{4}{*}{\rotatebox[origin=c]{90}{specialist}}} 
      & CUB200-2011 & 62.9 & 74.2 & 83.3 & 36.4 & 48.1 & 60.2 & 27.5 & 34.0 & 40.9 & 42.3  & 52.1 & 61.5 \\
      & CARS196 & 26.8 & 36.9 & 48.7 & 80.1 & 87.0 & 91.7 & 25.4 & 31.5 & 37.8 & 44.1 & 51.8 & 59.4 \\
      & In-shop & 9.2 & 13.9 & 20.5 & 16.4 & 21.3 & 27.7  & \textbf{85.1} & 90.2 & 93.6 & 36.9 & 41.8 & 47.3 \\
      & 3-way Concat. & 50.9 & 63.6 & 75.2 & 71.8 & 81.6 & 88.9 & 71.8 & 80.5 & 86.7 & 64.8 & 75.2 & 83.6 \\
    \midrule
    \parbox[t]{2mm}{\multirow{5}{*}{\rotatebox[origin=c]{90}{universal}}} 
    & Multi-Similarity~\cite{wang2019multi} & 52.9 & 66.3 & 76.7 & 74.3 & 83.5 & 89.8 & 84.5 & 90.2 & 93.8 & 70.6 & 80.0 & 86.8 \\
    & Multi-Similarity+DS & 59.0 & 70.9 & 80.5 & 78.4 & 86.3 & 91.6 & 83.3 & 89.3 & 93.4 & 73.6 & 82.2 & 88.5 \\
    & Multi-Similarity+BAL & 59.2 & 70.1 & 80.1 & 78.4 & 86.3 & 91.4 & 77.9 & 85.3 & 90.2 & 71.8 & 80.6 & 87.2 \\
    & Ours\_RKD & \textbf{65.3} & \textbf{76.3} & \textbf{84.5} & \textbf{83.8} & \textbf{90.3} & \textbf{93.8} & \textbf{85.1} & \textbf{90.3} & \textbf{93.9} & \textbf{78.1} & \textbf{85.6} & \textbf{90.7} \\
    & Ours\_SND & 63.9 & 75.4 & 84.2 & 80.7 & 88.2 & 92.8  & 84.6 & 89.8 & 93.5 & 76.4 & 84.5 & 90.2 \\
    \bottomrule
  \end{tabular}
  \end{adjustbox}
\vspace{-0.5cm}
\end{table}

To show that the proposed method has the capability to deal with specialists trained with other methods, we train CUB200-2011, CARS196, and In-shop specialists using one of the state-of-the-art methods, Multi-Similarity~\cite{wang2019multi}, and then distill the specialists into a universal embedding. The Recall@1 of the three specialists on single domain are $63.0$, $80.1$, and $85.4$, respectively. 
The performance of the unified models is listed in Table~\ref{tab:distill_ms}. \yf{The findings are very similar to what we obtain by distilling Triplet semi-hard models.}

\subsection{Unifying ImageNet with Other Domains}
We also try to unify ImageNet with CUB200-2011 or CARS196, which is quite different from the cases in Sec.~\ref{sec:exclusive}. The datasets used in Sec.~\ref{sec:exclusive} are much smaller than ImageNet and they are mutually exclusive. ImageNet contains $1,000$ categories over a broad range and it has a certain overlap with the small datasets. The total number of training images in ImageNet is 218 and 159 times larger than CUB200-2011 and CARS196, respectively. %
\yf{If we simply sample the mini-batch proportional to the number of training images, it is very rare to see the images from CUB200-2011 or CARS196. So we increase the probability of choosing the training images from CUB200-2011 or CARS196 by ten times. The result sampling policy can be viewed as a combination of domain-specific and domain-balance sampling and it is used for all of ``Triplet'', ``Ours\_RKD'', and ``Ours\_SND'' in this section.}
\yf{Since no distance shrinkage or early overfitting happens to the ImageNet dataset, it is not necessary to use SND for ImageNet. We directly use triplet loss instead of SND on the ImageNet images during universal embedding training. 
Because there exists category overlapping between ImageNet and small datasets, we do not fuse the evaluation set and the na\"ive sampling baseline cannot be used. The evaluation sets without fusion are adequate to evaluate the universal embedding in this section because the ImageNet evaluation set can show whether bird/car images are mixed with other images and the CUB200-2011/CARS196 evaluation set can show the fine-grained retrieval performance.}

\begin{table}[t]
  \setlength{\belowcaptionskip}{1pt}
  \caption{Performance comparisons of unifying ImageNet and CUB200-2011 in the left, and unifying ImageNet and CARS196 in the right. The recalls are computed without fusing the evaluation sets.}
  \label{tab:in_cub}
  \centering
  \begin{adjustbox}{max width=\columnwidth}
  \begin{tabular}{l|cccc|cccc||cccc|cccc}
    \toprule
    & \multicolumn{8}{c||}{Combination 3} & \multicolumn{8}{c}{Combination 4} \\
     & \multicolumn{4}{c|}{ImageNet} & \multicolumn{4}{c||}{CUB200-2011} & \multicolumn{4}{c|}{ImageNet} & \multicolumn{4}{c}{CARS196}\\
      & R1 & R2 & R4 & R8 & R1 & R2 & R4 & R8 & R1 & R2 & R4 & R8 & R1 & R2 & R4 & R8 \\
    \midrule
    ImageNet spec. & \textbf{65.3} & \textbf{74.0} & \textbf{80.2} & 84.6 & 44.6 & 55.8 & 68.0 & 79.1 & \textbf{65.3} & \textbf{74.0} & \textbf{80.2} & 84.6 & 27.6 & 38.5 & 50.7 & 63.5\\
    2-way Concat. & 63.8 & 73.1 & \textbf{80.2} & \textbf{85.5} & 58.3 & 70.9 & 80.8 & \textbf{88.7} & 61.5 & 71.8 & 79.4 & \textbf{85.0} & 66.2 & 77.3 & 85.5 & 91.2 \\
    \midrule
    Triplet~\cite{schroff2015facenet} & 64.6 & 73.4 & 79.8 & 84.6 & 51.3 & 63.1 & 74.5 & 83.1 & 64.7 & 73.5 & 80.0 & 84.7 & 51.5 & 64.3 & 75.0 & 83.4 \\
    Ours\_RKD & 64.5 & 73.5 & 79.7 & 84.5 & 59.2 & 70.9 & 80.6 & 87.9 & 64.1 & 73.1 & 79.7 & 84.4 & 69.5 & 79.3 & 86.5 & 91.7 \\
    Ours\_SND & 64.6 & 73.4 & 79.8 & 84.3 & \textbf{59.8} & \textbf{71.5} & \textbf{81.5} & 88.6 & 64.5 & 73.3 & 79.6 & 84.3 & \textbf{72.3} & \textbf{82.2} & \textbf{88.8} & \textbf{93.3} \\
    \bottomrule
  \end{tabular}
  \end{adjustbox}
\vspace{-0.5cm}
\end{table}

The results of the unification are reported in Table~\ref{tab:in_cub}. Compared with the specialists' performance in Table~\ref{tab:in_domain}, we find that the ImageNet specialist performs quite badly on CUB200-2011 and CARS196. As the bird and car categories in ImageNet are in a coarser granularity than in CUB200-2011 and CARS196, the ImageNet specialist does not know how to distinguish the fine-grained birds or cars. If we train a universal embedding using ``Triplet'' on the fused training data, the performance on CUB200-2011 and CARS196 are much better than ImageNet specialist. %
\yf{Although twice as much computation is used, the ``Concatenation'' baseline achieves inferior performance.} 
Again, the performance of the universal embedding trained with distillation is comparable to the performance of specialists. Different from Sec.~\ref{sec:exclusive}, the performance of ``Ours\_SND'' is better than ``Ours\_RKD'' in Table~\ref{tab:in_cub}, showing that ``Ours\_SND'' handles distance shrinkage better than ``Ours\_RKD''.

\section{Conclusions}
In this paper, we have studied the problem of how to train a universal image embedding model to have good performance on multiple domains. This problem is very important for large-scale image retrieval but has rarely been studied before. Fusing the training data from all the domains and training with single domain methods cannot solve this problem because of the early overfitting problem for some domains. To solve this problem, we propose to distill the knowledge learned by properly trained specialist models into a desired universal embedding model. When unifying a coarse-grained domain with a fine-grained domain, the learned knowledge, \emph{distances between images}, cannot be distilled directly because the distances are at different scales. Therefore, we develop a novel embedding knowledge distillation method based on SNE. The experimental results of unifying several combinations of public datasets have shown the effectiveness of the proposed method.
\bibliographystyle{splncs04}
\bibliography{egbib}
\end{document}